# Toward Supervised Anomaly Detection

**Nico Görnitz**                                                            NICO.GOERNITZ@TU-BERLIN.DE
**Marius Kloft**                                                            KLOFT@TU-BERLIN.DE
*Machine Learning Laboratory, Technische Universität Berlin*
*Franklinstr. 28/29, Berlin, Germany*
*Computational Biology Center*
*Memorial Sloan-Kettering Cancer Center*
*New York City, USA*

**Konrad Rieck**                                                            KONRAD.RIECK@UNI-GOETTINGEN.DE
*University of Göttingen, Dep. of Computer Science*
*Goldschmidtstr. 7, 37077 Göttingen, Germany*

**Ulf Brefeld**                                                             BREFELD@KMA.INFORMATIK.TU-DARMSTADT.DE
*Technische Universität Darmstadt*
*Hochschulstr. 10, 64289 Darmstadt, Germany*
*German Institute for International Educational Research*
*Schloßstr. 29, 60486 Frankfurt, Germany*

## Abstract

Anomaly detection is being regarded as an unsupervised learning task as anomalies stem from adversarial or unlikely events with unknown distributions. However, the predictive performance of purely unsupervised anomaly detection often fails to match the required detection rates in many tasks and there exists a need for labeled data to guide the model generation. Our first contribution shows that classical semi-supervised approaches, originating from a supervised classifier, are inappropriate and hardly detect new and unknown anomalies. We argue that semi-supervised anomaly detection needs to ground on the unsupervised learning paradigm and devise a novel algorithm that meets this requirement. Although being intrinsically non-convex, we further show that the optimization problem has a convex equivalent under relatively mild assumptions. Additionally, we propose an active learning strategy to automatically filter candidates for labeling. In an empirical study on network intrusion detection data, we observe that the proposed learning methodology requires much less labeled data than the state-of-the-art, while achieving higher detection accuracies.

## 1. Introduction

Anomaly detection deals with identifying unlikely and rare events. The classical approach to anomaly detection is to compute a precise description of normal data. Every newly arriving instance is contrasted with the model of normality and an anomaly score is computed. The score describes the deviations of the new instance compared to the average data instance and, if the deviation exceeds a predefined threshold, the instance is considered an anomaly or an outlier and processed adequately (Markou & Singh, 2003a; Chandola, Banerjee, & Kumar, 2009; Markou & Singh, 2003b).

Identifying data that exhibits irregular and suspicious traits is crucial in many applications such as medical imaging and network security. In particular, the latter has become a vivid research area as computer systems are increasingly exposed to security threats, such as computer worms, network attacks, and malicious code (Andrews & Pregibon, 1978). Network intrusion detection deals with detecting previously unknown threats and attacks in network traffic. Conventional security techniques for intrusion detection are based on identifying known patterns of misuse, so called *signatures* (Roesch, 1999; Paxson, 1999) and thus—although being effective against *known* attacks—fail to protect from novel threats. This brings anomaly detection into the focus of security research (e.g., Eskin, Arnold, Prerau, Portnoy, & Stolfo, 2002; Kruegel, Vigna, & Robertson, 2005; Stolfo, Apap, Eskin, Heller, Hershkop, Honig, & Svore, 2005; Perdisci, Ariu, Fogla, Giacinto, & Lee, 2009). Thus, anomaly detection is the most beneficial in learning scenarios where many regular data instances are given, which allows the machine to approximate the underlying distribution well and leads to a concise model of normality. By contrast, outliers and anomalies are rare and can even originate from changing distributions (e.g., novel classes of network attacks). Especially in adversarial settings, such as network intrusion detection, differences in training and test distributions are eminent as novel threats and tactics are being continuously developed. As a consequence, anomaly detection is generally considered an unsupervised task and prominent learning methods, including one-class support vector machines (Schölkopf, Platt, Shawe-Taylor, Smola, & Williamson, 2001) and support vector data descriptions (SVDD, Tax & Duin, 2004), implement this spirit. However, the underlying assumptions on unsupervised methods are also their major drawback and in many application areas, unsupervised methods fail to achieve the required detection rates. Especially in adversarial application areas such as network intrusion detection, even a single undetected outlier may already suffice to capture the system. Therefore, the goal of this article is to incorporate a feedback-loop in terms of labeled data to make anomaly detection practical. By doing so, knowledge about historic threats and anomalies can be included in terms of labels and thus guide the model generation toward better generalizations.

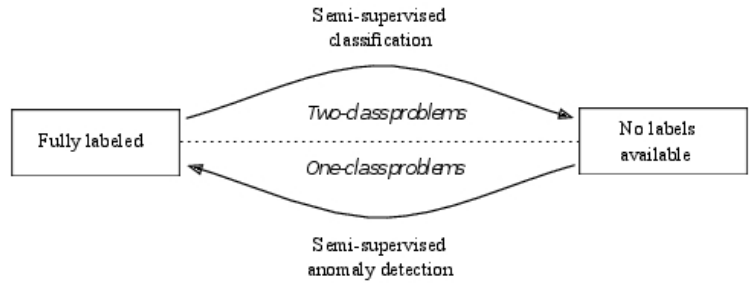

Figure 1: Illustration of the two paradigms for semi-supervised learning.

In this article, we cast anomaly detection into the paradigm of semi-supervised learning (Chapelle, Schölkopf, & Zien, 2006). Usually, semi-supervised methods are deduced from existing supervised techniques, augmented by an appropriate bias to take the unlabeled data into account. For instance, a prominent bias assumes that the unlabeled data is structured in clusters so that closeness (with respect to some measure) is proportional to the probability of having the same class label (Vapnik, 1998; Joachims, 1999; Chapelle & Zien, 2005; Chapelle, Chi, & Zien, 2006; Sindhwani, Niyogi, & Belkin, 2005). As a consequence, anomaly detection is often rephrased as a (multi-class) classification problem (Almgren & Jonsson, 2004; Stokes & Platt, 2008; Pelleg & Moore, 2004; Mao, Lee, Parikh, Chen, & Huang, 2009). Although assuming a cluster-structure of the data is often well justified in anomaly detection, recall that supervised learning techniques focus on discriminating concept classes while unsupervised techniques rather focus on data characterization. In this article, we show that differences in training and test distributions as well as the occurrence of previously unseen outlier classes render anomaly detection methods, derived from a supervised technique, inappropriate as they are likely to miss out novel and previously unseen classes of anomalies as depicted. By contrast, we argue that successful anomaly detection methods inherently need to ground on the unsupervised learning paradigm, see Figure 1. In sum, making anomaly detection practical requires the following key characteristics: (i) intrinsically following the unsupervised learning paradigm to cope with unknown events and (ii) additionally exploiting label information to obtain state-of-the-art results.

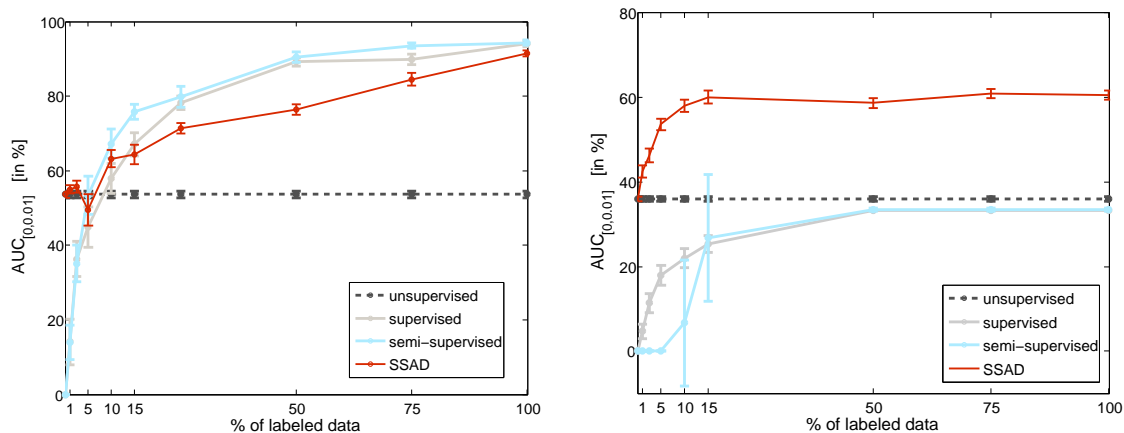

Figure 2: Left: The standard supervised classification scenario with identical training and test distributions. Right: The anomaly detection setting with two novel anomaly clusters in the test distribution.

In Figure 2, we show the results of a controlled experiment that visualizes the different nature of the semi-supervised methods derived from supervised and unsupervised paradigms, respectively. On the left hand side, the achieved accuracies in the standard supervised classification scenario, where training and test distributions are identical, is shown. The performance of the unsupervised anomaly detection method is clearly outperformed by supervised and semi-supervised approaches. However, we observe from the right hand side of the figure how fragile the latter methods can be in an anomaly detection scenario. The experimental setup is identical to the former except that we discard two anomaly clusters in the training set (see Figure 3). Note that this change is not an arbitrary modification but an inherent characteristic of anomaly detection scenarios, where anomalies stem from novel and previously unseen distributions. Unsurprisingly, the (partially) supervised methods fail to detect the novel outliers and are clearly outperformed by the unsupervised approach, which robustly performs around unimpressive detection rates of about 40%. Finally, all methods are clearly outperformed by our novel semi-supervised anomaly detection (SSAD) which is devised from the unsupervised learning paradigm and allows for incorporating labeled data.

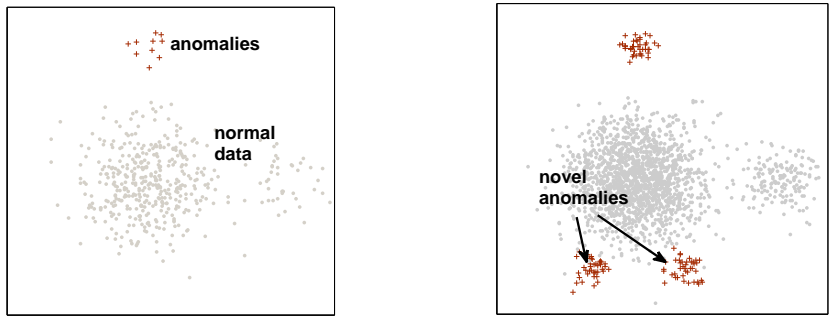

Figure 3: Left: training data stems from two clusters of normal data (gray) and one small anomaly cluster (red). Right: two additional anomaly clusters (red) appear in the test data set.

The main contribution of this article is to provide a mathematical sound methodology for semi-supervised anomaly detection. Carefully conducted experiments and their discussions show the importance of distinguishing between the two semi-supervised settings as depicted in Figure 1. To meet this requirement, we propose a novel semi-supervised anomaly detection technique that is derived from the unsupervised learning paradigm, but allows for incorporating labeled data in the training process. Our approach is based on the support vector data description (SVDD) and contains the original formulation as a special case. Although the final optimization problem is not convex, we show that an equivalent convex formulation can be obtained under relatively mild assumptions. To guide the user in the labeling process, we additionally propose an active learning strategy to improve the actual model and to quickly detect novel anomaly clusters. We empirically evaluate our method on network intrusion detection tasks. Our contribution proves robust in scenarios where the performance of baseline approaches deteriorate due to obfuscation techniques. In addition, the active learning strategy is shown to be useful as a standalone method for threshold adaptation.

The remainder of this article is organized as follows. Section 2 reviews related work. The novel semi-supervised anomaly detection methods are presented in Section 3 and Section 4 introduces active learning strategies. Section 5 gives insights into the proposed learning paradigm and we report on results for real-world network intrusion scenarios in Section 6. Section 7 concludes.

## 2. Related Work

Semi-supervised learning (Chapelle, Schölkopf et al., 2006) offers a mathematical sound framework for learning with partially labeled data. For instance, transductive approaches to semi-supervised learning assume a cluster structure in the data so that close points are likely to share the same label while points that are far away are likely to be labeled differently. The transductive support vector machine (TSVM, Vapnik, 1998; Joachims, 1999; Chapelle & Zien, 2005) optimizes a max-margin hyperplane in feature space that implements the cluster assumption. In its most basic formulation, the TSVM is a non-convex integer programming problem on top of an SVM. This is computationally very expensive so that Chapelle, Chi et al. (2006) propose a more efficient smooth relaxation of the TSVM. Another related approach is low density separation (LDS) by Chapelle and Zien (2005), where the cluster structure is modeled by a graph of distances.

A broad overview of anomaly detection can be found in the work of Chandola et al. (2009). Anomaly detection is being regarded as unsupervised learning task and therefore it is not surprising that there exist a large number of applications employing unsupervised anomaly detection methods. For instance, finding anomalies in network traffic (Eskin et al., 2002) or program behaviour (Heller, Svore, Keromytis, & Stolfo, 2003), denoising patterns (Park, Kang, Kim, Kwok, & Tsang, 2007) or annotating (Goh, Chang, & Li, 2005) and classifying images (Lai, Tax, Duin, Zbieta, Ekalska, & Ik, 2004) and documents (Manevitz & Yousef, 2002; Onoda, Murata, & Yamada, 2006)

Fully-supervised approaches for anomaly detection usually ignore unlabeled data during the training-phase: for example, Almgren and Jonsson (2004) employ a max-margin classifier that separates the innocuous data from the attacks. Stokes and Platt (2008) present a technique which combines approaches for effective discrimination (Almgren & Jonsson, 2004) and rare-class detection (Pelleg & Moore, 2004). Mao et al. (2009) take a multi-view and co-training approach based on Blum and Mitchell (1998) to learn from labeled and unlabeled data.

Support vector learning has also been extended to many non-standard settings such as one-class learning (Schölkopf et al., 2001) and support vector data description (Tax & Duin, 2004). The idea of the latter is to learn a hypersphere that encloses the bulk of the provided data so that all instances that lie outside of the hypersphere are considered anomalous. By contrast, the one-class SVM learns a hyperplane in some feature space that divides the data points from the origin with maximum-margin. For translation-invariant kernel matrices, both approaches are equivalent.

There exist only a few semi-supervised methods that are based on unsupervised techniques. Blanchard, Lee, and Scott (2010) propose a method which has the appealing option of specifying an upper threshold on the false-positives rate. However, this method needs to include test instances at training time and is not applicable in online and streaming scenarios such as anomaly detection. The same holds true for an extension of the one-class SVM by Mũnoz Marí, Bovolo, Gómez-Chova, Bruzzone, and Camp-Valls (2010) that incorporates labeled examples in a graph-Laplacian regularization term. Tax (2001) proposes a straight-forward extension of the SVDD to semi-supervised anomaly detection, where negatively labeled points are required to lie outside of the hypersphere—otherwise a penalty is incurred. An advantage of this so-called $\text{SVDD}^{neg}$ approach is that no further assumptions on the underlying data-generating probability distribution such as manifold assumptions are imposed. Unfortunately, the primal $\text{SVDD}^{neg}$ problem is not a convex optimization problem, which makes it very difficult to accurately optimize. Moreover, dual optimization as proposed in the work of Tax (2001) cannot be considered an sound alternative due to possible duality gaps. However, in Appendix A we show that there is a convex reformulation of the $\text{SVDD}^{neg}$ for translation-invariant kernels, such as RBF-kernels. The new formulation does not suffer from duality gaps and can be easily solved by primal or dual descent methods. The same problem occurs in related semi-supervised one-class methods as proposed by Liu and Zheng (2006) and Wang, Neskovic, and Cooper (2005).

Another broad class of methods deals with learning from positive and unlabeled examples (LPUE). Intrinsically, one aims at solving a two-class problem but only data from one class (the positive class) is given together with unlabeled data points. LPUE can thus be applied to the problem setting at hand by identifying the outlier class with positively labeled data. Zhang and Lee (2005) show that this class of methods can be viewed as a special case of semi-supervised learning and emphasize that the $\text{SVDD}^{neg}$ (Tax, 2001) can be considered an instance of LPUE. Algorithmically, LPUE is often solved in an iterative manner by (i) identifying a reliable set of labeled examples using a classifier and (ii) re-training the classifier given the new training set (Liu, Dai,

Li, Lee, & Yu, 2003; Zhang & Lee, 2005; Blum & Mitchell, 1998). Though some work addresses learning from non-i.i.d. data (e.g., Li & Liu, 2005), the underlying assumption usually implies that training and test sets are drawn from the same distribution.

The present article builds upon a previous paper of the same authors (Görnitz, Kloft, & Brefeld, 2009). It extends the latter by a mathematical sound framework and intuitive philosophical insights. In addition, we present a more general problem formulation employing arbitrary convex loss functions and the computation of the dual representation thereof, and a new empirical analysis with comparisons to a larger variety of baseline approaches.

## 3. Semi-supervised Anomaly Detection

In anomaly detection tasks, we are given $n$ observations $\bm{x}_1, \ldots, \bm{x}_n \in \mathcal{X}$. The underlying assumption is that the bulk of the data stems from the same (unknown) distribution and we call this part of the data *normal*. Some few observations, however, originate from different distributions and are considered *anomalies*. These anomalies could for instance be caused by broken sensors or network attacks and cannot be sampled by definition. The goal in anomaly detection is to detect these anomalies by finding a concise description of the normal data, so that deviating observations become outliers. We thus aim at finding a scoring function $f : \mathcal{X} \to \mathcal{R}$ which defines the model of normality. Following the principle of empirical risk minimization, the optimization problem takes the following form,

$$f^* = \underset{f}{\operatorname{argmin}}\, \Omega(f) + \frac{\eta}{n} \sum_{i=1}^{n} l(f(\bm{x}_i)),$$

where $l : \mathcal{R} \to \mathcal{R}$ is an appropriate loss function, $\Omega : \mathcal{R}^d \to \mathcal{R}^+$ a regularizer on $f$, and $\eta$ is a trade-off parameter.

Our approach is based on the SVDD, which computes a hypersphere with radius $R$ and center $\bm{c}$ that encompasses the data. The hypersphere is our model of normality and the anomaly score for an instance $\bm{x}$ is computed by its distance to the center $\bm{c}$,

$$f(\bm{x}) = ||\phi(\bm{x}) - \bm{c}||^2 - R^2. \tag{1}$$

Points lying outside of the ball (i.e., $f(\bm{x}) > 0$) are considered anomalous, while points within ($f(\bm{x}) < 0$) are treated as normal data. The corresponding optimization problem is known as support vector data description (SVDD, Tax, 2001) and has the following form

$$\begin{aligned}
\min_{R, \bm{c}, \bm{\xi}} \quad & R^2 + \eta_u \sum_{i=1}^{n} \xi_i \\
\text{s.t.} \quad & \forall_{i=1}^{n} : \|\phi(\bm{x}_i) - \bm{c}\|^2 \leq R^2 + \xi_i \\
& \forall_{i=1}^{n} : \xi_i \geq 0,
\end{aligned}$$

where the trade-off $\eta_u$ balances the minimization of the radius and the sum of erroneously placed points (that are, points lying outside of the normality radius). The parameter $\eta_u$ also serves as an estimate of the ratio between outliers and normal data in the $n$ training examples. The resulting problem is convex and can be solved equivalently in dual space using the representation $\bm{c} = \sum_{i=1}^{n} \alpha_i \phi(\bm{x}_i)$. As a consequence, the input data can be expressed equivalently by a kernel function $k(\bm{x}_i, \bm{x}_j) = \phi(\bm{x}_i)^T \phi(\bm{x}_j)$ on $\mathcal{X}$ that corresponds to a feature map $\phi : \mathcal{X} \to \mathcal{F}$ into a reproducing kernel Hilbert space $\mathcal{F}$ (see, e.g., Müller, Mika, Rätsch, Tsuda, & Schölkopf, 2001).

### 3.1 Semi-supervised Anomaly Detection

We now propose a novel approach to semi-supervised anomaly detection. The proposed method generalizes the vanilla SVDD and processes unlabeled *and* labeled examples. While existing extensions of the SVDD employ dual optimization techniques and inherently suffer from duality gaps due to their non-convexity, we propose a primal approach to semi-supervised anomaly detection. As discussed earlier, for translation-invariant kernels, the one-class SVM is contained in our framework as a special case. In addition to the $n$ unlabeled examples $x_1, \ldots, x_n \in \mathcal{X}$, we are now given $m$ labeled observations $(x_1^*, y_1^*), \ldots, (x_m^*, y_m^*) \in \mathcal{X} \times \mathcal{Y}$ where $Y$ denotes the set of class labels. For simplicity, we will focus on $Y = \{+1, -1\}$ where $y^* = +1$ encodes nominal data and $y^* = -1$ anomalies.

As argued in the introduction, the goal is to derive a method that grounds on the unsupervised learning paradigm. We therefore stick to the hypersphere model of the SVDD and use the latter as blueprint for dealing with unlabeled data. The inclusion of labeled examples follows a simple pattern: If an example $x^*$ is labeled as nomial ($y^* = +1$), we require that it lies within the hypersphere. By contrast, if an example is an anomaly or member of an outlier class ($y^* = -1$), we want to have it placed outside of the ball. A straight-forward extension of the SVDD using both labeled and unlabeled examples is thus given by

$$
\begin{aligned}
\min_{R, \gamma, c, \xi} \quad & R^2 - \kappa\gamma + \eta_u \sum_{i=1}^{n} \xi_i + \eta_l \sum_{j=n+1}^{n+m} \xi_j^* \\
\text{s.t.} \quad & \forall_{i=1}^{n} : \quad \|\phi(x_i) - c\|^2 \leq R^2 + \xi_i \\
& \forall_{j=n+1}^{n+m} : \quad y_j^* \left( \|\phi(x_j^*) - c\|^2 - R^2 \right) \leq -\gamma + \xi_j^* \\
& \forall_{i=1}^{n} : \quad \xi_i \geq 0, \\
& \forall_{j=n+1}^{n+m} : \quad \xi_j^* \geq 0,
\end{aligned}
\tag{2}
$$

where $\gamma$ is the margin of the labeled examples and $\kappa$, $\eta_u$, and $\eta_l$ are trade-off parameters. Unfortunately, the inclusion of negatively labeled data renders the above optimization problem non-convex and optimization in dual space is prohibitive. As a remedy, following the approach of Chapelle and Zien (2005), we translate Equation (2) into an unconstrained problem. We thereby resolve the slack terms from the above OP as follows

$$
\begin{aligned}
\xi_i &= \ell \left( R^2 - \|\phi(x_i) - c\|^2 \right) \\
\xi_j^* &= \ell \left( y_j^* \left( R^2 - \|\phi(x_j^*) - c\|^2 \right) - \gamma \right).
\end{aligned}
\tag{3}
$$

For example, if we put $\ell(t) = \max\{-t, 0\}$ (i.e., the common hinge loss), we recover (2). Furthermore, by an application of the representer theorem, we obtain the support-vector expansion

$$
c = \sum_{i=1}^{n} \alpha_i \phi(x_i) + \sum_{j=n+1}^{n+m} \alpha_j y_j^* \phi(x_j^*)
\tag{4}
$$

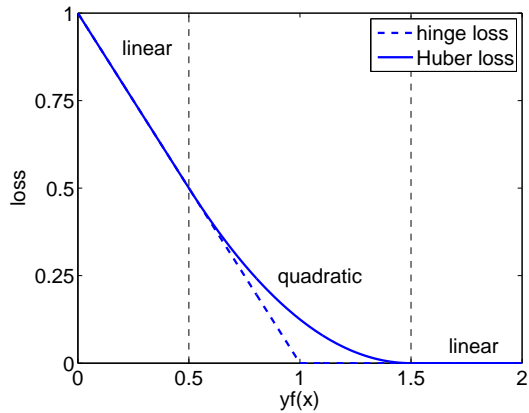

Figure 4: Non-differentiable hinge loss (dashed) and differentiable Huber loss $\ell_{\Delta=1,\epsilon=0.5}$ (solid).

(see Appendix B for a detailed derivation). Combining (3) and (4), we can re-formulate optimization problem (2) solely in terms of kernels and *without* any constraints as follows:

$$\min_{R,\gamma,\boldsymbol{\alpha}} \quad R^2 - \kappa\gamma + \eta_u \sum_{i=1}^{n} \ell_\epsilon \left( R^2 - k(\boldsymbol{x}_i, \boldsymbol{x}_i) + (2\boldsymbol{e}_i - \boldsymbol{\alpha})' K \boldsymbol{\alpha} \right)$$
$$+ \eta_l \sum_{j=n+1}^{n+m} \ell_\epsilon \left( y_j^* \left( R^2 - k(\boldsymbol{x}_j^*, \boldsymbol{x}_j^*) + (2\boldsymbol{e}_j^* - \boldsymbol{\alpha})' K \boldsymbol{\alpha} \right) - \gamma \right). \quad (5)$$

Hereby $K = (k_{ij})_{1 \le i,j \le n}$ denotes the kernel matrix given by $k_{ij} = k(\boldsymbol{x}_i, \boldsymbol{x}_j) = \langle \phi(\boldsymbol{x}_i), \phi(\boldsymbol{x}_j) \rangle$ and $\boldsymbol{e}_1, \ldots, \boldsymbol{e}_{n+m}$ is the standard base of $\mathbb{R}^{n+m}$. By rephrasing the problem as an unconstrained optimization problem, its intrinsic complexity has not changed. Often, unconstrained optimization is easier to implement than constrained optimization. While non-smooth optimization is possible via e.g. non-convex bundle methods as described in the work of Do (2010), smooth optimization methods such as conjugate gradient or Newton's method are easier to apply. To obtain a smooth optimization technique, we choose *Huber's robust loss* (Huber, 1972). The Huber loss has two parameters controlling its quadratic approximation in terms of the center $\Delta$ and its witdh $\epsilon$, see Figure 4. The optimization function becomes differentiable and off-the-shelf gradient-based optimization tools can be applied. The complete derivation of the gradients of optimization problem (5) using Huber's robust loss is shown in Appendix C.

### 3.2 Convex Semi-supervised Anomaly Detection

The optimization problem of the previous section is easy to implement but, unfortunately, non-convex. Therefore, optimizers may find a good local optimum, but several restarts are necessary to verify the quality of the solutions and in cases optimization might fail completely. We now show that, under rather mild assumptions, namely that the data is processed to have unit norm in feature space (as fulfilled by, e.g., RBF kernels), the above optimization problem can be converted into an equivalent convex one. Our derivation is very general as it postulates nothing but the convexity of the loss function. Our approach is based on a combination of Lagrangian duality and the notion of the Fenchel-Legendre conjugate function. Fenchel duality for machine learning has been pioneered by Rifkin and Lippert (2007) under the assumption of full-rank kernels. Our approach is more general and allows us to use *any* kernel $K$. As a byproduct of our derivation, we show that the classical one-class SVM is a special case of a general class of density level set estimators that minimize a convex risk functional and give a general dual criterion for this class. To this aim, we introduce the Legendre-Fenchel conjugate for a given loss $l(t)$ as

$$l_c(z) = \sup_t \left(zt - l(t)\right)$$

and use a slightly different formulation of the SSAD problem, that is, we eliminate the hinge loss slack variables $\boldsymbol{\xi}^*, \boldsymbol{\xi}$ and reformulate the problem with explicit loss functions:

$$\min_{\rho,\gamma,\boldsymbol{w},\boldsymbol{t}} \quad \frac{1}{2}||\boldsymbol{w}||^2 - \rho - \kappa\gamma + \eta_u \sum_{i=1}^n l(t_i) + \eta_l \sum_{j=n+1}^{n+m} l(t_j)$$
$$\text{s.t.} \quad \forall_{i=1}^n : \quad t_i = (\boldsymbol{w}^T \phi(\boldsymbol{x}_i)) - \rho \qquad \text{(P)}$$
$$\forall_{j=n+1}^{n+m} : \quad t_j = (y_j^* \boldsymbol{w}^T \phi(\boldsymbol{x}_j^*)) - y_j^* \rho - \gamma$$
$$\text{and } \gamma \geq 0.$$

Note that, in the above problem, auxiliary variables $t_i$ are introduced to deal with non-differentiable loss functions. Again, because of the convex nature of the stated optimization problem, we can solve it in the dual space. To this aim, we use the Lagrange Theorem to incorporate the constraints into the objective:

$$\begin{aligned}L =& \frac{1}{2}||\boldsymbol{w}||^2 - \rho - \kappa\gamma + \eta_u \sum_{i=1}^n l(t_i) + \eta_l \sum_{j=n+1}^{n+m} l(t_j) \\ & - \sum_{i=1}^n \alpha_i((\boldsymbol{w}^T \phi(\boldsymbol{x}_i)) - \rho - t_i) \\ & - \sum_{j=n+1}^{n+m} \alpha_j((y_j^* \boldsymbol{w}^T \phi(\boldsymbol{x}_j^*)) - y_j^* \rho - \gamma - t_j) - \delta\gamma\end{aligned} \qquad (6)$$

An optimal solution can be found by solving the Lagrangian saddle point problem

$$\max_{\boldsymbol{\alpha},\delta} \min_{\rho,\gamma,\boldsymbol{w},\boldsymbol{t}} EQ6.$$

If we used a standard Lagrangian ansatz, we would now compute the derivate of the Lagrangian with respect to the primal variables. However, a general loss function $l(\cdot)$ is not necessarily differentiable. As a remedy, we only compute the derivatives wrt $\boldsymbol{w}$, $\rho$ and $\gamma$. Setting those to zero, yields the optimality conditions

$$\begin{aligned}\forall i : & \quad 0 \leq \alpha_i \leq \eta_u \\ \forall j : & \quad 0 \leq \alpha_j \leq \eta_l \\ \boldsymbol{w} & = \sum_{i=1}^n \alpha_i \phi(\boldsymbol{x}_i) + \sum_{j=n+1}^{n+m} \alpha_j y_j^* \phi(\boldsymbol{x}_j^*).\end{aligned} \qquad (7)$$

Inserting the above optimality conditions into the Lagrangian, the saddle point problem translates into

$$\max_{\boldsymbol{\alpha}} -\frac{1}{2}\boldsymbol{\alpha}^T K\boldsymbol{\alpha} + \eta_u \sum_{i=1}^{n} \min_{\boldsymbol{t}} \left( l(t_i) + \frac{\alpha_i}{\eta_u} t_i \right) + \eta_l \sum_{j=n+1}^{n+m} \min_{\boldsymbol{t}^*} \left( l(t_j^*) + \frac{\alpha_j}{\eta_l} t_j^* \right).$$

Converting the min into a max statement results in

$$\max_{\boldsymbol{\alpha}} -\frac{1}{2}\boldsymbol{\alpha}^T K\boldsymbol{\alpha} - \eta_u \sum_{i=1}^{n} \max_{\boldsymbol{t}} \left( -\frac{\alpha_i}{\eta_u} t_i - l(t_i) \right) - \eta_l \sum_{j=n+1}^{n+m} \max_{\boldsymbol{t}^*} \left( -\frac{\alpha_j}{\eta_l} t_j^* - l(t_j^*) \right).$$

Now, making use of the Legendre-Fenchel conjugate $l_c(\cdot)$ described above, we arrive at the following dual optimization problem

$$\max_{\boldsymbol{\alpha}} \quad -\frac{1}{2}\boldsymbol{\alpha}^T K\boldsymbol{\alpha} - \eta_u \sum_{i=1}^{n} l_c(-\frac{\alpha_i}{\eta_u}) - \eta_l \sum_{j=n+1}^{n+m} l_c(-\frac{\alpha_j}{\eta_l}) \quad (D)$$

$$\text{s.t.} \quad 1 = \sum_{i=1}^{n} \alpha_i + \sum_{j=n+1}^{n+m} \alpha_j y_j^* \quad \text{and} \quad \kappa \leq \sum_{j=n+1}^{n+m} \alpha_j.$$

In contrast to existing semi-supervised approaches to anomaly detection (Tax, 2001; Liu & Zheng, 2006; Wang et al., 2005), strong duality holds as shown by the following proposition.

**Proposition 3.1** *For the optimization problems (P) and (D) strong duality holds.*

**Proof** This follows from the convexity of (P) and Slater's condition, which is trivially fulfilled for all $\gamma$ by adjusting $t_j$: $\forall \gamma > 0 \quad \exists t_j \in \mathbb{R} : 0 = (y_j^* \boldsymbol{w}^T \phi(\boldsymbol{x}_j^*)) - y_j^* \rho - \gamma - t_j$. □

We observe that the above optimization problems (P) and (D) contain the non-convex variant as a special case for translation-invariant kernels. Difficulties may arise in the presence of many equality and inequality constraints, which can increase computational requirements. However, this is not inherent; the left hand-side constraint can be removed by discarding the variable $\rho$ in the initial primal problem—this leaves the regularization path of the optimization problem invariant—and the right hand side inequality can be equivalently incorporated into the objective function by a Lagrangian argument (e.g., Proposition 12 in Kloft, Brefeld, Sonnenburg, & Zien, 2011). Note, that the convex model has only an intuitive interpretation for normalized kernels. In order to deal with a wider class of kernels, we need to resort to the more general non-convex formulation as presented in Section 3.1.

## 4. Active Learning for Semi-supervised Anomaly Detection

In the previous section, we presented two optimization problems that incorporate labeled data into an unsupervised anomaly detection technique. However, we have not yet addressed the question of acquiring labeled examples. Many topical real-world applications involve millions of training instances (Sonnenburg, 2008) so that domain experts can only label a small fraction of the unlabeled data. *Active learning* deals with finding the instances that, once labeled and included in the training set, lead to the largest improvement of a re-trained model. In the following, we present an active learning strategy that is well-suited for anomaly detection. The core idea is to query low-confidence decisions to guide the user in the labeling process.

Our approach works as follows. First, we initialize our method by training it on the unlabeled examples. The training set is then augmented by particular examples that have been selected by the active learning rule. The candidates are labeled by a domain expert and added to the training set. The model is retrained on the refined training set, which now consists of unlabeled *and* labeled examples. Subsequently labeling- and retraining-steps are repeated until the required performance is reached.

The active learning rule itself consists of two parts. We begin with a commonly used active learning strategy which simply queries borderline points. The idea of the method is to choose the point that is closest to the decision hypersphere (Almgren & Jonsson, 2004; Warmuth, Liao, Rätsch, Mathieson, Putta, & Lemmen, 2003) to be presented to the expert:

$$ \boldsymbol{x}' \;=\; \operatorname*{argmin}_{\boldsymbol{x}\in\{\boldsymbol{x}_1,\ldots,\boldsymbol{x}_n\}} \frac{\|f(\boldsymbol{x})\|}{\max_k \|f(\boldsymbol{x}_k)\|} \;=\; \operatorname*{argmin}_{\boldsymbol{x}\in\{\boldsymbol{x}_1,\ldots,\boldsymbol{x}_n\}} \left\| R^2 - \|\phi(\boldsymbol{x}) - \boldsymbol{c}\|^2 \right\|. \tag{8} $$

For supervised support vector machines, this strategy is known as the *margin strategy* (Tong & Koller, 2000). Figure 5 (a) shows an illustratation for semi-supervised anomaly detection.

When dealing with non-stationary outlier categories, it is beneficial to identify *novel* anomaly classes as soon as possible. We translate this requirement into an active learning strategy as follows. Let $A = (a_{ij})_{i,j=1,\ldots,n+m}$ be an adjacency matrix of the training instances, obtained by, for example, a $k$-nearest-neighbor approach, where $a_{ij} = 1$ if $\boldsymbol{x}_i$ is among the $k$-nearest neighbors of $\boldsymbol{x}_j$ and 0 otherwise. We introduce an extended labeling $\bar{y}_1 \ldots, \bar{y}_{n+m}$ for all examples by defining $\bar{y}_i = 0$ for unlabeled instances and retaining the labels for labeled instances, i.e., $\bar{y}_j = y_j$. Using these pseudo labels, Equation (9) returns the unlabeled instance according to

$$ \boldsymbol{x}' \;=\; \operatorname*{argmin}_{\boldsymbol{x}_i\in\{\boldsymbol{x}_1,\ldots,\boldsymbol{x}_n\}} \frac{1}{2k} \sum_{j=1}^{n+m} (\bar{y}_j + 1)\, a_{ij}. \tag{9} $$

The above strategy explores unknown clusters in feature space and thus labels orthogonal or complementary instances as illustrated in Figure 5 (b).

Nevertheless, using Equation (9) alone may result in querying points lying close to the center of the actual hypersphere. These points will hardly contribute to an improvement of the hypersphere. On the other hand, using the margin strategy alone does not allow for querying novel regions that lie far away from the margin. In other words, only a combination of both strategies (8) and (9) guarantees that points of interest are queried. Our final active learning strategy is therefore given by

$$ \boldsymbol{x}' = \operatorname*{argmin}_{\boldsymbol{x}_i\in\{\boldsymbol{x}_1,\ldots,\boldsymbol{x}_n\}} = \delta\frac{\|f(\boldsymbol{x})\|}{c} + \frac{1-\delta}{2k} \sum_{j=1}^{n+m} (\bar{y}_j + 1)\, a_{ij} \tag{10} $$

for $\delta \in [0, 1]$. The combined strategy queries instances that are close to the boundary of the hypersphere *and* lie in potentially anomalous clusters with respect to the $k$-nearest neighbor graph, see Figure 5 (c) for an illustration. Depending on the actual value of $\delta$, the strategy jumps from cluster to cluster and thus helps to identify interesting regions in feature space. For the special case of no labeled points, our combined strategy reduces to the margin strategy.

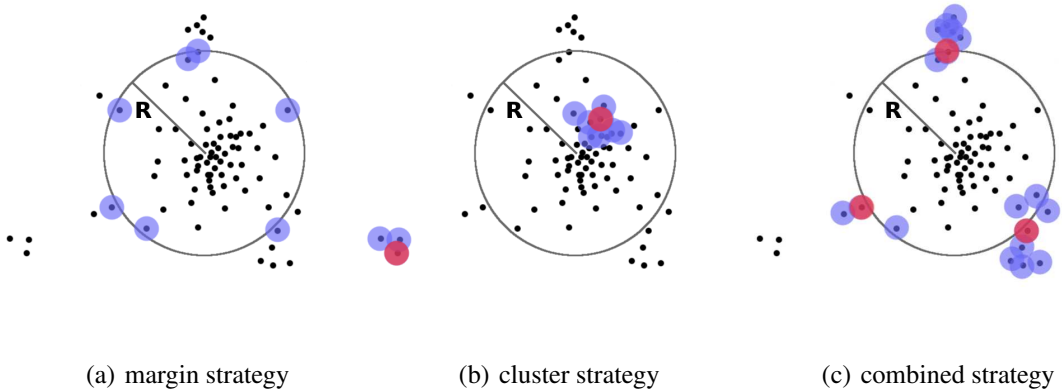

(a) margin strategy  (b) cluster strategy  (c) combined strategy

Figure 5: Comparison of active learning strategies (queried points are marked in blue): (a) the margin strategy queries data points that are closest to the decision boundary, (b) the cluster strategy queries points in rarely labeled regions, and (c) the combined strategy queries data points that are likely anomalies in clusters near the decision boundary.

Usually, an active learning step is followed by an optimization step of the semi-supervised SVDD, to update the model with respect to recently labeled data. This procedure is of course time-consuming and can be altered for practical settings, for instance by querying a couple of points before performing a model update. Irrespectively of the actual implementation, alternating between active learning and updating the model can be repeated until a desired predictive performance is obtained.

## 5. Illustration of Proposed Learning Paradigm

In this section, we illustrate the weaknesses of existing learning paradigms in semi-supervised anomaly detection settings by means of a controlled experiment on synthetic data. The results, which have already been briefly sketched in the introduction (cf., Figure 2), are discussed in detail here.

Table 1: Competitors for the toy data experiment.

| two-class | | one-class | | |
| --- | --- | --- | --- | --- |
| supervised | transductive | unsupervised | LPUE | semi-supervised |
| SVM | LDS | SVDD | $SVDD^{neg}$ | SSAD |

To this end, we generate the nominal training and validation data from two isotropic Gaussian distributions in $\mathbb{R}^2$ and one anomaly cluster (shown in Figure 3 (left)). However, at testing time, two novel anomaly clusters appear in test data (shown in Figure 3 (right)). This reflects the characteristic that anomalies can stem from novel, previously unseen distributions. We compare our newly-developed method SSAD to the following baseline approaches: the unsupervised support vector domain description (SVDD, Tax & Duin, 2004), the corrected semi-supervised SVDD$^{neg}$ (Tax, 2001) described in Appendix A, a supervised support vector machine (SVM, Boser, Guyon, & Vapnik, 1992; Cortes & Vapnik, 1995), and the semi-supervised low-density separation (LDS, Chapelle & Zien, 2005), see Table 1. As common in anomaly detection setups, we measure the area under the ROC curve over the interval $[0, 0.01]$ and report on AUCs averaged over 25 repetitions with distinct training, validation, and test sets. In every repetition, parameters $\eta_u$, $\eta_l$ are adjusted on the respective validation set within the interval $[10^{-2}, 10^2]$. In all experiments, we used $\kappa = 1$. Error bars correspond to standard errors.

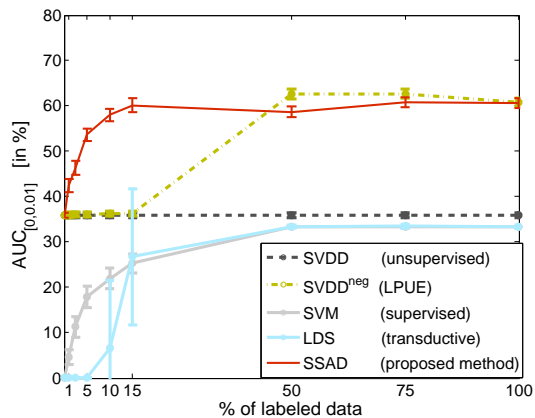

Figure 6: Performance of various unsupervised, supervised and semi-supervised methods in the anomaly detection setting.

The results are shown in Figure 6, where the horizontal axis shows different ratios of labeled and randomly drawn unlabeled examples. Methods derived from the supervised learning paradigm such as SVM and LDS cannot cope with novel outlier clusters and perform poorly for all ratios of labeled and unlabeled examples; their performance remains below that of the unsupervised SVDD, which does not utilize labeled data at all and is thus unaffected by incorporating labels in the training process. By contrast, the two semi-supervised methods derived from the unsupervised learning paradigm clearly outperform all other baselines. However, the SVDD$^{neg}$ only benefits from anomalous labeled data and since these are sparse, needs a big fraction of labeled data to increase its performance. Our semi-supervised method SSAD exploits every single labeled example and needs only 15% of the labels to saturate around its optimum.

Figure 7 visualizes typical contour lines of hypotheses computed by SSAD for three different scenarios. The figure shows a fully-supervised scenario where all instances are correctly labeled (left), a semi-supervised solution where 25% of the data is labeled and 75% remains unlabeled (center), and a completely unsupervised one using unlabeled data only (right). Unsurprisingly, the fully supervised solution discriminates perfectly between normal data and outliers while the unsupervised solution recognizes the latter as normal data by mistake. The intermediate semi-supervised solution uses only little label information to also achieve a perfect separation of the involved classes.

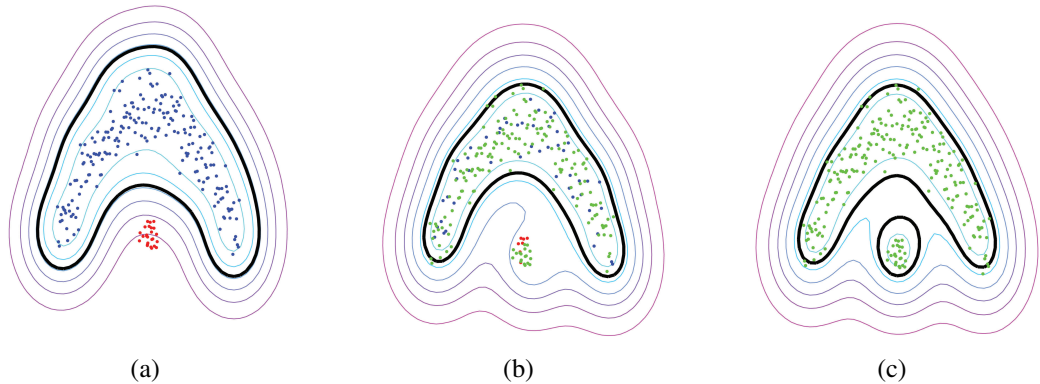

Figure 7: Different solutions for fully-supervised (left), semi-supervised (center), and unsupervised anomaly detection using RBF kernels. Colors indicate unlabeled data (green), labeled outliers (red), and labeled normal instances (violet).

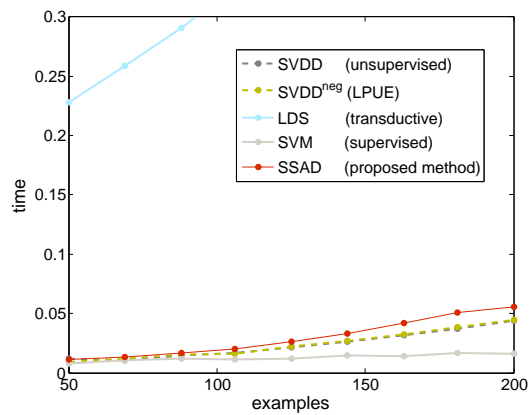

Figure 8: Execution times.

Figure 8 compares execution times of the different methods and shows the number of training examples versus training time. For simplicity, we discarded 50% of the labels in the training data at random. The results show that methods, such as SSAD, SVDD$^{neg}$, and SVDD, that are derived from the unsupervised learning principle, perform similarly. The SVM performs best but uses only the labeled part of the training data and ignores the unlabeled examples. Low density separation (LDS) performs worst due to its transductive nature.

Based on our observations, we draw the following conclusions. Anomaly detection scenarios render methods derived from the supervised learning paradigm inappropriate. Even unsupervised methods ignoring label information may perform better than their supervised peers. Intuitively, discarding label information is sub-optimal. Our experiment shows that semi-supervised methods from the unsupervised learning paradigm effectively incorporate label information and outperform all other competitors. We will confirm these findings in Section 6.

## 6. Real-World Network Intrusion Detection

The goal of network intrusion detection is to identify attacks in incoming network traffic. Classical signature-based have proven insufficient for the identification of *novel* attacks, because signatures need to be manually crafted in advance. Therefore machine learning approaches have been gaining more and more attention by the intrusion detection research community.

The detection of unknown and novel attacks requires an adequate representation of network contents. In the remainder, we apply a technique for embedding network payloads in vector spaces derived from concepts of information retrieval (Salton, Wong, & Yang, 1975) and that has recently been applied in the application domain of network intrusion detection (Rieck & Laskov, 2007). A network payload $x$ (the data contained in a network packet or connection) is mapped to a vector space using a set of strings $S$ and an embedding function $\phi$. For each string $s \in S$ the function $\phi_s(x)$ returns 1 if $s$ is contained in the payload $x$ and 0 otherwise. By applying $\phi_s(x)$ for all elements of $S$ we obtain the following map

$$\phi : \mathcal{X} \to \mathbb{R}^{|S|}, \quad \phi : x \mapsto (\phi_s(x))_{s \in S}, \tag{11}$$

where $\mathcal{X}$ is the domain of all network payloads. Defining a set $S$ of relevant strings *a priori* is difficult as typical patterns of novel attacks are not available prior to their disclosure. As an alternative, we define the set $S$ implicitly and associate $S$ with all possible strings of length $n$. This resulting set of strings is often referred to as $n$-grams.

As a consequence of using $n$-grams, the network payloads are mapped to a vector space with $256^n$ dimensions, which apparently contradicts efficient network intrusion detection. Fortunately, a payload of length $T$ comprises at most $(T - n)$ different $n$-grams and, consequently, the map $\phi$ is *sparse*, that is, the vast majority of dimensions is zero. This sparsity can be exploited to derive linear-time algorithms for extraction and comparison of embedded vectors. Instead of operating with full vectors, only non-zero dimensions are considered, where the extracted strings associated with each dimension can be maintained in efficient data structures (Rieck & Laskov, 2008).

For our experiments, we consider HTTP traffic recorded within 10 days at Fraunhofer Institute FIRST. The data set comprises 145,069 unmodified connections of average length of 489 bytes. The incoming byte stream of each connection is mapped to a vector space using 3-grams as detailed above. We refer to the FIRST data as the *normal pool*. The *malicious pool* contains 27 real attack classes generated using the Metasploit framework (Maynor, Mookhey, Cervini, & Beaver, 2007). It covers 15 buffer overflows, 8 code injections and 4 other attacks including HTTP tunnels and cross-site scripting. Every attack is recorded in 2–6 different variants using a virtual network environment and a decoy HTTP server, where the attack payload is adapted to match characteristics of the normal data pool. A detailed description of this data set is provided by Rieck (2009).

To study the robustness of our approach in a more realistic scenario, we also consider techniques to obfuscate malicious content by adapting attack payloads to mimic benign traffic in feature space (Fogla, Sharif, Perdisci, Kolesnikov, & Lee, 2006; Perdisci et al., 2009). As a consequence, the extracted features deviate less from normality and the classifier is likely to be fooled by the attack. For our purposes, it already suffices to study a simple cloaking technique by adding common HTTP headers to the payload while the malicious body of the attack remains unaltered. We apply this technique to the malicious pool and refer to the obfuscated set of attacks as *cloaked pool*.

### 6.1 Detection Performance

In this section, we evaluate the statistical performance of SSAD in intrusion detection, in comparison to the baseline methods SVDD and SVDD$^{neg}$. In addition, our combined active learning strategy is compared to random sampling.

We focus on two scenarios: normal vs. malicious and normal vs. cloaked data. For both settings, the respective byte streams are translated into a bag-of-3-grams representation. For each experiment, we randomly draw 966 training examples from the normal pool and 34 attacks, depending on the scenario, either from the malicious or the cloaked pool. Holdout and test sets are also drawn at random and consist of 795 normal connections and 27 attacks, each. We make sure that attacks of the same attack class occur either in the training, or in the test set but not in both. We report on 10 repetitions with distinct training, holdout, and test sets and measure the performance by the area under the ROC curve in the false-positive interval $[0, 0.01]$ (AUC$_{0.01}$)

Figure 9(a) shows the results for normal vs. malicious data pools, where the x-axis depicts the percentage of randomly drawn labeled instances. Irrespectively of the amount of labeled data, the malicious traffic is detected by all methods equally well as the intrinsic nature of the attacks is well captured by the bag-of-3-grams representation (cf., Wang, Parekh, & Stolfo, 2006; Rieck & Laskov, 2006). There is no significant difference between the classifiers.

Figure 9(b) shows the results for normal vs. cloaked data. First of all, the performance of the unsupervised SVDD drops to just 70%. We obtain a similar result for the SVDD$^{neg}$; incorporating cloaked attack information into the training process of the SVDD leads to an increase of about 5% which is far from any practical value. Notice that the SVDD$^{neg}$ cannot make use of labeled data of the normal class. Thus, its moderate ascent in terms of the number of labeled examples is credited to the class ratio of $966/34$ for the random labeling strategy. The bulk of additional information cannot be exploited and has to be left out. By contrast, our semi-supervised method SSAD includes all labeled data into the training process and clearly outperforms the two baselines. For only 5% labeled data, SSAD easily beats the best baseline and for randomly labeling 15% of the available data it separates normal and cloaked malicious traffic almost perfectly.

Nevertheless, labeling 15% of the data is not realistic for practical applications. We thus explore the benefit of active learning for inquiring label information of borderline and low-confidence points. Figure 9(c) shows the results for normal vs. cloaked data, where the labeled data for SVDD$^{neg}$ and SSAD is chosen according to the active learning strategy in Equation (10). The unsupervised SVDD does not make use of labeled information and is unaffected by this setup, remaining at an AUC$_{0.01}$ of 70%. Compared to the results when using a random labeling strategy (Figure 9(b)), the performance of the SVDD$^{neg}$ increases significantly. The ascent of the SVDD$^{neg}$ is now steeper and its performance yields 85% for just 15% labeled data. However, SSAD also improves for active learning and dominates the baselines. Using active learning, we need to label only 3% of the data for attaining an almost perfect separation, compared to 25% for a random labeling strategy. We conclude that our active learning strategy effectively improves the performance and reduces the manual labeling effort significantly.

In Figure 10 the impact of our active learning strategy given by Equation (10) is shown. We compare the number of outliers detected by the combined strategy with the margin-based strategy in Equation (8) (see also, Almgren & Jonsson, 2004) and by randomly drawing instances from the unlabeled pool. As a sanity check, we also included the theoretical outcome for random sampling.

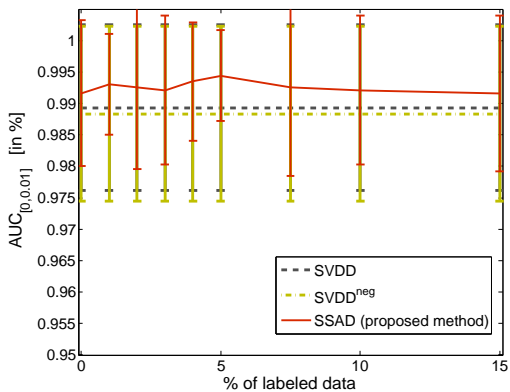
(a) Detection accuracies of regular attacks.

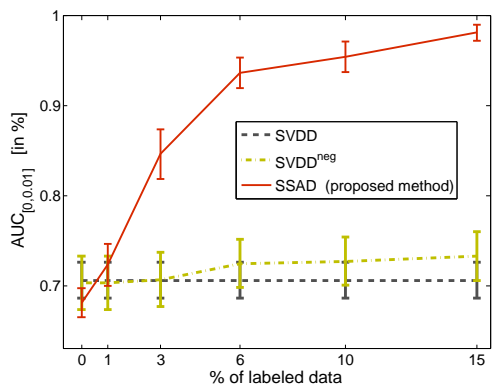
(b) Detection accuracies of cloaked attacks.

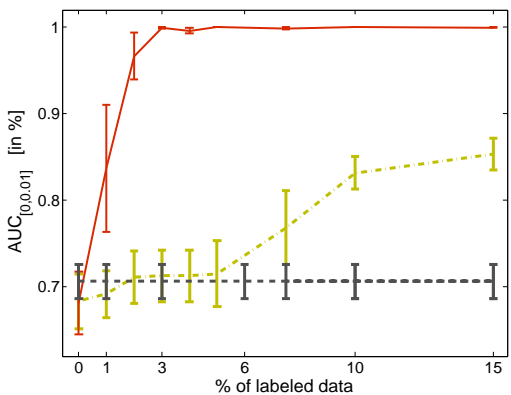
(c) Detection accuracies of cloaked attacks using proposed active learning strategy for SSAD.

Figure 9: *Results of the network intrusion detection experiment.* While detection accuracies of regular attacks are insignificantly different (see (a)), SSAD achieves up to 30% higher accuracies than baseline approaches for cloaked data (see (b)). The proposed activity learning strategy further increases the accuracy when labeled data is rare (see (c)).

The results show that the combined strategy effectively detects malicious traffic much faster than the margin-based strategy.

### 6.2 Threshold Adaptation

The previous experiments have demonstrated the advantages of active learning for network intrusion detection. So far, all results have been obtained using our method SSAD; however, the active learning techniques devised in Section 4 are also applicable for calibrating other learning-based methods. We herein focus on the vanilla SVDD with parameter value $\nu = 1$, which corresponds to classical centroid-based anomaly detection (e.g., Shawe-Taylor & Cristianini, 2004), such that results directly transfer to anomaly detectors as Anagram (Wang et al., 2006), McPad (Perdisci et al., 2009) and TokDoc (Krueger, Gehl, Rieck, & Laskov, 2010).

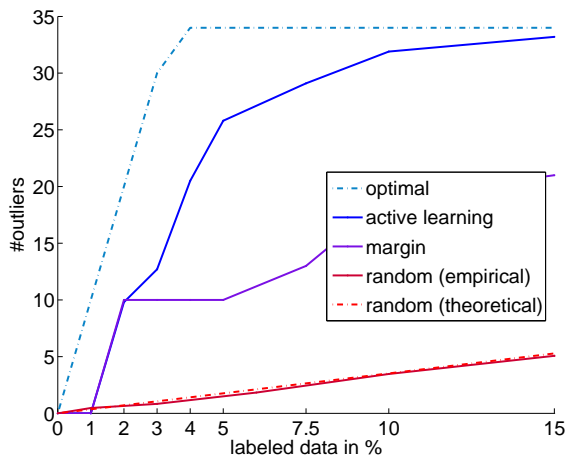

Figure 10: Number of novel attacks detected by the combined active learning strategy (blue line), random sampling (red solid and dotted line), margin strategy (purple line) and upper bound (light blue dotted line) for a single run.

We again draw a set of 3,750 network connections from the pool of normal data and split the resulting set into a training set of 2,500 connections and a test partition of 1,250 events. Both sets are mixed with cloaked attack instances. The SVDD is then trained on the normal training set delivering a threshold $R$. For application of the learned hypersphere to the test set, we evaluate different strategies for determining a radius $\hat{R}$ using random sampling and active learning. In both cases, the selected connections are labeled and a threshold is obtained by computing the mean of all labeled instances:

$$\hat{R} = \begin{cases} R & : \quad \#pos = 0 \quad \wedge \quad \#neg = 0 \\ \max_i d(\boldsymbol{x}_i) & : \quad \#pos > 0 \quad \wedge \quad \#neg = 0 \\ \min_j d(\boldsymbol{x}_j) & : \quad \#pos = 0 \quad \wedge \quad \#neg > 0 \\ \frac{\sum_i d(\boldsymbol{x}_i) + \sum_j d(\boldsymbol{x}_j)}{\#pos + \#neg} & : \quad \#pos > 0 \quad \wedge \quad \#neg > 0 \end{cases} \quad (12)$$

Where $\boldsymbol{x}_i$ are the positive labeled examples, $\boldsymbol{x}_j$ the negative examples and $d(\boldsymbol{x}) = ||\phi(\boldsymbol{x}) - \boldsymbol{c}||$ denotes the distance of the current sample $\boldsymbol{x}$ from the hyperspheres origin. Figure 11 shows the ROC curve of the SVDD and the computed thresholds for various levels of labeled data. Results have been averaged over 10 random draws of working sets. One can see that even for small amounts of labeled data the active learning strategy finds a reasonable radius while the random strategy and the vanilla SVDD completely fail with a false-positive rate of 0.5 and 1, respectively. This result demonstrates that active learning strategies enable calibrating anomaly detectors with a significantly reduced effort in comparison to random sampling and hence provide a valuable instrument when deploying learning methods in practice.

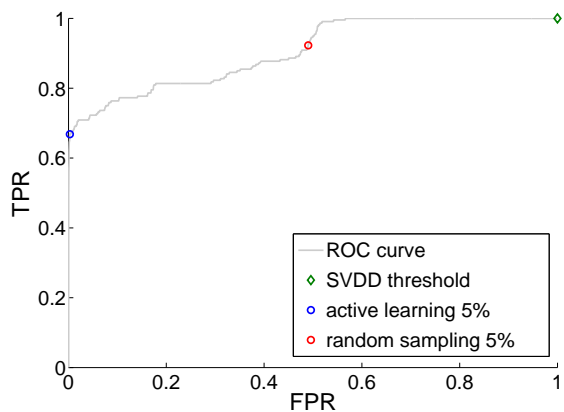

Figure 11: Results of the threshold adaption experiment: ROC curve of the SVDD (grey) and thresholds as determined by the SVDD (green), our proposed combined strategy (blue), and random sampling (red) are shown (labeling 5% of the data).

## 7. Conclusion

In this article, we developed a framework for semi-supervised anomaly detection, which allows for the inclusion of prior and expert knowledge. We discussed the conceptual difference of semi-supervised models which are derived from unsupervised or supervised techniques and proposed a generalization of the support vector data description to incorporate labeled data. The optimization problem of semi-supervised anomaly detection (SSAD) is an unconstrained, continuous problem, that allows for an efficient optimization by gradient-based methods and has a convex equivalent under mild assumptions on the kernel function.

We approached semi-supervised anomaly detection from an unsupervised learning paradigm. We proposed a novel active learning strategy that is specially tailored to anomaly detection. Our strategy guides the user to in the labeling process by querying instances that are not only close to the boundary of the hypersphere, but are also likely to contain instances of novel outlier categories.

Empirically, we applied semi-supervised anomaly detection to the application domain of network intrusion detection. We showed that rephrasing the unsupervised problem as a semi-supervised task is beneficial in practice: SSAD proves robust in scenarios where the performance of baseline approaches deteriorates due to obfuscation techniques. Moreover, we demonstrated the effectiveness of our active learning strategy on a couple of data sets and observed SSAD to significantly improve the prediction accuracy by effectively exploiting the limited amount of labeled available. We observed that only a handful labeled instances are necessary to boost the performance. This characteristic is especially appealing in tasks where labeling data is costly such as network security where the traffic has to be inspected for malicious patterns by an expert expert.

There are many possibilities to exploit and extend our learning approach as well as our active learning strategy. For example, replacing the $\ell_2$-norm regularization by a sparsity-inducing $\ell_1$-norm to incorporate automatic feature selection reduces the dimensionality of the solution. Learning sparse feature representations are of great interest for other computer security applications such as signature generation. A possible optimization strategy could be the linear programming (LP) approach by Campbell and Bennett (2001) for data domain description. However, other choices of regularizers are certainly possible including structured regularizers to incorporate hierarchies in the learning process or non-isotropic norms to encode additional domain knowledge. Incorporating multiple labels and rephrasing semi-supervised anomaly detection as a multi-task problem might also improve accuracy in complex application domains.

## Acknowledgments

The authors are very grateful to Klaus-Robert Müller for comments that helped improving the manuscript. This work was supported in part by the German Bundesministerium für Bildung und Forschung (BMBF) under the project PROSEC (FKZ 01BY1145), by the FP7-ICT Programme of the European Community, under the PASCAL2 Network of Excellence, and by the German National Science Foundation (DFG) under GA 1615/1-1, MU 987/6-1, MU 987/11-1 and RA 1894/1-1. Furthermore, Marius Kloft acknowledges a PhD scholarship by the German Academic Exchange Service (DAAD) and a postdoctoral fellowship by the German Research Foundation (DFG) as well as funding by the Ministry of Education, Science, and Technology, through the National Research Foundation of Korea under Grant R31-10008. A part of the work was done while Marius Kloft was with Computer Science Division and Department of Statistics, University of California, Berkeley, CA 94720-1758, USA.

## Appendix A. Analysis of SVDD$^{neg}$

In this appendix, we point out a limitation of previously published methods such as SVDD$^{neg}$ (Tax, 2001) as well as methods proposed by Hoi, Chan, Huang, Lyu, and King (2003), Liu and Zheng (2006), Wang et al. (2005), and Yuan and Casasent (2004). These methods suffer from potential duality gaps as they are optimized in dual space but—depending on the training data—run the risk of originating from a non-convex optimization problem. For instance, this is the case if only a single negatively labeled example is included in the training set. This issue is not addressed in the aforementioned papers.

We exemplarily illustrate the problem for the SVDD$^{neg}$. The SVDD$^{neg}$ incorporates labeled examples of the outlier class into the otherwise unsupervised learning process. As before, the majority of the (unlabeled) data points shall lie inside the sphere while the labeled outliers are constrained to lie outside of the normality ball. This results in a two-class problem where the positive class consists of unlabeled data and the negative class is formed by the labeled outliers. After introducing class labels $y \in \{+1, -1\}$ (where unlabeled data points receive the class label $y_i = +1$), the primal optimization problem is given by

$$\min_{R, \boldsymbol{c}, \boldsymbol{\xi}} \quad R^2 + \eta \sum_{i=1}^{n} \xi_i$$
$$\text{s.t.} \quad \forall_{i=1}^{n} : y_i \|\phi(\boldsymbol{x}_i) - \boldsymbol{c}\|^2 \leq y_i R^2 + \xi_i \quad \text{and} \quad \xi_i \geq 0, \tag{13}$$

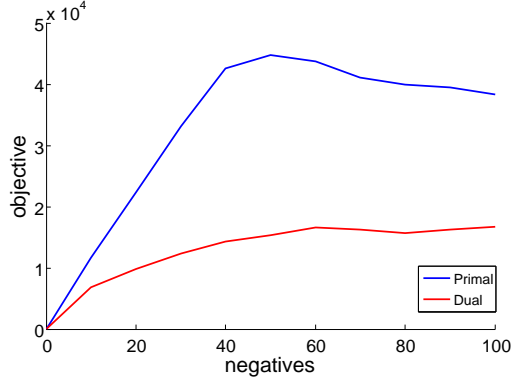

Figure 12: Exemplary duality gap for SVDD$^{neg}$ using a linear kernel and $\eta = 100$. The horizontal axis shows the percentage of negative (anomalous) points in the training set and the vertical axis shows the primal and dual objective values.

and the corresponding optimization problem in dual space is given by

$$\max_{\boldsymbol{\alpha}} \quad \sum_{i=1}^{n} \alpha_i y_i k(\boldsymbol{x}_i, \boldsymbol{x}_i) - \sum_{i,j=1}^{n} \alpha_i \alpha_j y_i y_j k(\boldsymbol{x}_i, \boldsymbol{x}_j) \qquad (14)$$

$$\text{s.t.} \quad \sum_{i=1}^{n} \alpha_i y_i = 1 \quad \text{and} \quad 0 \leq \alpha_i \leq \eta \quad \forall i = 1, \ldots, n.$$

The existence of the duality gap is shown as follows: The second derivative of the primal constraints $g(\boldsymbol{x}_i) = y_i \|\phi(\boldsymbol{x}_i) - \boldsymbol{c}\|^2 - y_i R^2 - \xi_i \leq 0$ given by $\partial^2 g / \partial \boldsymbol{c}^2 = 2y_i$ is negative for outliers as their label equals $y_i = -1$. This turns the whole optimization problem non-convex. As a consequence, the optimal solutions of the primal and dual problems may differ. Figure 12 shows an exemplary plot of the duality gap for artificially generated data where one nominal Gaussian is surrounded by a smaller 'anomalous' Gaussian. During the labeling process more and more data points receive their corresponding label and the more negative examples are present in the learning problem (horizontal axis) the larger is the duality gap and the larger is the difference of the two objective values (vertical axis). Note that the duality gap is not necessarily a monotonic function although the behavior is likely the case. Furthermore, the maximization of the dual problem yields a lower bound on the primal objective (blue line), whereas the latter is always greater or equal than the corresponding dual (red line).

Nevertheless, the following Theorem shows for the class of translation-invariant kernel functions, there exists an equivalent convex re-formulation in form of the one-class SVM (Schölkopf et al., 2001).

**Theorem A.1** *The solution $\boldsymbol{\alpha}^*$ found by optimizing the dual of the non-convex SVDD$^{neg}$ as stated in Equations* (14) *is identical to the dual of the corresponding convex one-class SVM problem as stated in Equation* (15) *if the kernel is translation-invariant, i.e., $k(\boldsymbol{x}_i, \boldsymbol{x}_i) = s \quad \forall i, s \in \mathbb{R}^+$.*

**Proof** The dual of the one-class SVM is given by

$$\max_{\boldsymbol{\alpha}} \quad -\frac{1}{2}\sum_{i,j=1}^{n}\alpha_i\alpha_j y_i y_j k(\boldsymbol{x}_i,\boldsymbol{x}_j), \qquad \text{s.t.} \quad \sum_{i=1}^{n}\alpha_i y_i = 1 \quad \text{and} \quad 0 \leq \alpha_i \leq \eta \quad \forall i. \tag{15}$$

The respective constraints are already equivalent. The dual SVDD$^{neg}$ objective with translation-invariant kernel reduces to

$$\begin{aligned}
\boldsymbol{\alpha}^* &= \operatorname*{argmax}_{\boldsymbol{\alpha}} \quad \sum_{i=1}^{n}\alpha_i y_i k(\boldsymbol{x}_i,\boldsymbol{x}_i) - \sum_{i,j=1}^{n}\alpha_i\alpha_j y_i y_j k(\boldsymbol{x}_i,\boldsymbol{x}_j) \\
&= \operatorname*{argmax}_{\boldsymbol{\alpha}} \quad ns\sum_{i=1}^{n}\alpha_i y_i - \sum_{i,j=1}^{n}\alpha_i\alpha_j y_i y_j k(\boldsymbol{x}_i,\boldsymbol{x}_j) \\
&= \operatorname*{argmax}_{\boldsymbol{\alpha}} \quad ns - \sum_{i,j=1}^{n}\alpha_i\alpha_j y_i y_j k(\boldsymbol{x}_i,\boldsymbol{x}_j),
\end{aligned} \tag{16}$$

where we substituted the equality constraint of Eq. (15) in the last step. Finally, Equation (16) is precisely the one-class SVM dual objective, scaled by $\frac{1}{2}$ and shifted by a constant $ns$. However, the optimal solution $\boldsymbol{\alpha}^*$ is not affected by this transformation which completes the proof. $\square$

## Appendix B. A Representer Theorem for SSAD

In this section, we show the applicability of the representer theorem for semi-supervised anomaly detection.

**Theorem B.1 (Representer Theorem in Schölkopf & Smola, 2002)** *Let $\mathcal{H}$ be a reproducing kernel Hilbert space with a kernel $k : \mathcal{X} \times \mathcal{X} \to \mathbb{R}$, a symmetric positive semi-definite function on the compact domain. For any function $L : \mathcal{R}^n \to \mathbb{R}$, any nondecreasing function $\Omega : \mathbb{R} \to \mathbb{R}$. If*

$$J^* := \min J(f)_{f\in\mathcal{H}} := \min f \in \mathcal{H}\{\Omega\left(||f||_{\mathcal{H}}^2\right) + L\left(f(\boldsymbol{x}_1),\ldots,f(\boldsymbol{x}_n)\right)\}$$

*is well-defined, then there exist $\alpha_1,\ldots,\alpha_n \in \mathbb{R}$, such that*

$$f(\cdot) = \sum_{i=1}^{n}\alpha_i k(\boldsymbol{x}_i,\cdot) \tag{17}$$

*achieves $J(f) = J^*$. Furthermore, if $\Omega$ is increasing, then each minimizer of $J(f)$ can be expressed in the form of Eq. (17).*

**Proposition B.2** *The representer theorem can be applied to the non-expanded version of Equation (5).*

**Proof** Recall the primal SSAD objective function which is given by

$$\begin{aligned}
J(R,\gamma,\boldsymbol{c}) = &R^2 - \kappa\gamma + \eta_u\sum_{i=1}^{n}\ell\left(R^2 - ||\phi(\boldsymbol{x}_i) - \boldsymbol{c}||^2\right) \\
&+ \eta_l\sum_{j=n+1}^{n+m}\ell\left(y_j^*\left(R^2 - ||\phi(\boldsymbol{x}_j^*) - \boldsymbol{c}||^2\right) - \gamma\right).
\end{aligned}$$

Substituting $T := R^2 - ||c||^2$ leads to the new objective function

$$J(T, \gamma, c) = ||c||^2 + T - \kappa\gamma + \eta_u \sum_{i=1}^{n} \ell \left(T - ||\phi(x_i)||^2 + 2\phi(x_i)'c\right)$$
$$+ \eta_l \sum_{j=n+1}^{n+m} \ell \left(y_j^* \left(T - ||\phi(x_j^*)||^2 + 2\phi(x_j^*)'c\right) - \gamma\right).$$

Expanding the center $c$ in terms of labeled and unlabeled input examples is now covered by the representer theorem. After the optimization, $T$ can be easily re-substituted to obtain the primal variables $R$, $\gamma$, and $c$. This completes the proof. $\square$

## Appendix C. Computing the Gradients for Eq. (5)

In this section we compute the gradients of the SSAD formulation given by Eq. (5). This is a neccessary step to implement the gradient-based solver for SSAD. To this end, we consider the unconstrained optimization problem given by

$$\min_{R, \gamma, \alpha} \quad R^2 - \kappa\gamma + \eta_u \sum_{i=1}^{n} \ell_{\Delta, \epsilon} \left(R^2 - k(x_i, x_i) + (2e_i - \alpha)' K \alpha\right)$$
$$+ \eta_l \sum_{j=n+1}^{n+m} \ell_{\Delta, \epsilon} \left(y_j^* \left(R^2 - k(x_j^*, x_j^*) + (2e_j^* - \alpha)' K \alpha\right) - \gamma\right),$$

where $\ell_{\Delta, \epsilon}$ is the Huber loss given by

$$\ell_{\Delta, \epsilon}(t) = \begin{cases} \Delta - t & : \quad t \leq \Delta - \epsilon \\ \frac{(\Delta + \epsilon - t)^2}{4\epsilon} & : \quad \Delta - \epsilon \leq t \leq \Delta + \epsilon \\ 0 & : \quad \text{otherwise.} \end{cases}$$

For notational convenience, we focus on the Huber loss for $\ell_{\Delta=0, \epsilon}(t)$ and move margin dependent terms into the argument $t$ and compute the gradients in several steps, as follows: first, we build the gradient with respect to the primal variables $R$ and $c$, which yields

$$\frac{\partial \xi_i}{\partial R} = 2R\ell'_\epsilon(R^2 - ||\phi(x_i) - c||^2)$$
$$\frac{\partial \xi_i}{\partial c} = 2(\phi(x_i) - c)\ell'_\epsilon(R^2 - ||\phi(x_i) - c||^2). \quad (18)$$

The derivatives of their counterparts $\xi_j^*$ for the labeled examples with respect to $R$, $\gamma$, and $c$ are given by

$$\frac{\partial \xi_j^*}{\partial R} = 2y_j^* R \ell'_\epsilon \left(y_j^* \left(R^2 - ||\phi(x_j^*) - c||^2\right) - \gamma\right)$$
$$\frac{\partial \xi_j^*}{\partial \gamma} = -\ell'_\epsilon \left(y_j^* \left(R^2 - ||\phi(x_j^*) - c||^2\right) - \gamma\right)$$
$$\frac{\partial \xi_j^*}{\partial c} = 2y_j^*(\phi(x_j^*) - c)\ell'_\epsilon \left(y_j^* \left(R^2 - ||\phi(x_j^*) - c||^2\right) - \gamma\right).$$

Substituting the partial gradients, we resolve the gradient of Equation (5) with respect to the primal variables as follows:

$$\frac{\partial EQ5}{\partial R} = 2R + \eta_u \sum_{i=1}^{n} \frac{\partial \xi_i}{\partial R} + \eta_l \sum_{j=n+1}^{n+m} \frac{\partial \xi_j^*}{\partial R}, \qquad (19)$$

$$\frac{\partial EQ5}{\partial \gamma} = -\kappa + \eta_l \sum_{j=n+1}^{n+m} \frac{\partial \xi_j^*}{\partial \gamma}, \qquad (20)$$

$$\frac{\partial EQ5}{\partial \boldsymbol{c}} = \eta_u \sum_{i=1}^{n} \frac{\partial \xi_i}{\partial \boldsymbol{c}} + \eta_l \sum_{j=n+1}^{n+m} \frac{\partial \xi_j^*}{\partial \boldsymbol{c}}. \qquad (21)$$

In the following, we extend our approach to allow for the use of kernel functions. An application of the representer theorem shows that the center $\boldsymbol{c}$ can be expanded as

$$\boldsymbol{c} = \sum_{i=1}^{n} \alpha_i \phi(\boldsymbol{x}_i) + \sum_{j=n+1}^{n+m} \alpha_j y_j^* \phi(\boldsymbol{x}_j^*). \qquad (22)$$

According to the chain rule, the gradient of Equation (5) with respect to the $\alpha_{i/j}$ is given by

$$\frac{\partial EQ5}{\partial \alpha_{i/j}} = \frac{\partial EQ5}{\partial \boldsymbol{c}} \frac{\partial \boldsymbol{c}}{\partial \alpha_{i/j}}.$$

Using Equation (22), the partial derivatives $\frac{\partial \boldsymbol{c}}{\partial \alpha_{i/j}}$ resolve to

$$\frac{\partial \boldsymbol{c}}{\partial \alpha_i} = \phi(\boldsymbol{x}_i) \quad \text{and} \quad \frac{\partial \boldsymbol{c}}{\partial \alpha_j} = y_j^* \phi(\boldsymbol{x}_j^*), \qquad (23)$$

respectively. Applying the chain-rule to Equations (19),(20),(21), and (23) gives the gradients of Equation (5) with respect to the $\alpha_{i/j}$.

# References

Almgren, M., & Jonsson, E. (2004). Using active learning in intrusion detection. In *Proc. of IEEE Computer Security Foundation Workshop*, pp. 88–89.

Andrews, D. F., & Pregibon, D. (1978). Finding the outliers that matter. *Journal of the Royal Statistical Society. Series B (Methodological)*, *40*(1), 85–93. /

Blanchard, G., Lee, G., & Scott, C. (2010). Semi-Supervised Novelty Detection. *Journal of Machine Learning Research*, ", 2973–2973–3009–3009.

Blum, A., & Mitchell, T. (1998). Combining labeled and unlabeled data with co-training. In *COLT' 98: Proc. of the eleventh annual conference on Computational learning theory*, pp. 92–100, New York, NY, USA. ACM.

Boser, B., Guyon, I., & Vapnik, V. (1992). A training algorithm for optimal margin classifiers. In Haussler, D. (Ed.), *Proceedings of the 5th Annual ACM Workshop on Computational Learning Theory*, pp. 144–152.

Toward Supervised Anomaly Detection

Campbell, C., & Bennett, K. (2001). A linear programming approach to novelty detection. In Leen, T., Dietterich, T., & Tresp, V. (Eds.), *Advances in Neural Information Processing Systems*, Vol. 13, pp. 395–401. MIT Press.

Chandola, V., Banerjee, A., & Kumar, V. (2009). Anomaly detection: A survey. *ACM Computing Surveys*, *41*(3), 1–58.

Chapelle, O., Chi, M., & Zien, A. (2006). A continuation method for semi-supervised SVMs. In *ICML*, pp. 185–192, New York, New York, USA. ACM.

Chapelle, O., & Zien, A. (2005). Semi-supervised classification by low density separation. In *Proc. of the International Workshop on AI and Statistics*.

Chapelle, O., Schölkopf, B., & Zien, A. (2006). *Semi-Supervised Learning (Adaptive Computation and Machine Learning)*. MIT Press.

Cortes, C., & Vapnik, V. (1995). Support vector networks. *Machine Learning*, *20*, 273–297.

Do, T.-M.-T. (2010). *Regularized bundle methods for large-scale learning problems with an application to large margin training of hidden Markov models*. Ph.D. thesis, Pierre and Marie Curie University Paris.

Eskin, E., Arnold, A., Prerau, M., Portnoy, L., & Stolfo, S. (2002). *Applications of Data Mining in Computer Security*, chap. A geometric framework for unsupervised anomaly detection: detecting intrusions in unlabeled data. Kluwer.

Fogla, P., Sharif, M., Perdisci, R., Kolesnikov, O., & Lee, W. (2006). Polymorphic blending attacks. In *Proc. of USENIX Security Symposium*.

Goh, K.-S., Chang, E. Y., & Li, B. (2005). Using one-class and two-class svms for multiclass image annotation. *IEEE Transactions on Knowledge and Data Engineering*, *17*, 1333–1346.

Görnitz, N., Kloft, M., & Brefeld, U. (2009). Active and semi-supervised data domain description. In *ECML/PKDD (1)*, pp. 407–422.

Heller, K., Svore, K., Keromytis, A., & Stolfo, S. (2003). One class support vector machines for detecting anomalous windows registry accesses. In *Proc. of the workshop on Data Mining for Computer Security*.

Hoi, C.-H., Chan, C.-H., Huang, K., Lyu, M., & King, I. (2003). Support vector machines for class representation and discrimination. In *Proc. of the International Joint Conference on Neural Networks*.

Huber, P. (1972). Robust statistics: a review. *Ann. Statist.*, *43*, 1041.

Joachims, T. (1999). Transductive inference for text classification using support vector machines. In *International Conference on Machine Learning (ICML)*, pp. 200–209, Bled, Slowenien.

Kloft, M., Brefeld, U., Sonnenburg, S., & Zien, A. (2011). $\ell_p$-norm multiple kernel learning. *Journal of Machine Learning Research*, *12*, 953–997.

Kruegel, C., Vigna, G., & Robertson, W. (2005). A multi-model approach to the detection of web-based attacks. *Computer Networks*, *48*(5).

Krueger, T., Gehl, C., Rieck, K., & Laskov, P. (2010). TokDoc: A self-healing web application firewall. In *Proc. of 25th ACM Symposium on Applied Computing (SAC)*, pp. 1846–1853.

Lai, C., Tax, D. M. J., Duin, R. P. W., Zbieta, E., Ekalska, P., & Ik, P. P. (2004). A study on combining image representations for image classification and retrieval. *International Journal of Pattern Recognition and Artificial Intelligence*, *18*, 867–890.

Li, X.-l., & Liu, B. (2005). Learning from Positive and Unlabeled Examples with Different Data Distributions. In *ECML*.

Liu, B., Dai, Y., Li, X., Lee, W. S., & Yu, P. S. (2003). Building Text Classifiers Using Positive and Unlabeled Examples. In *IEEE International Conference on Data Mining*, pp. 179–186. IEEE Comput. Soc.

Liu, Y., & Zheng, Y. F. (2006). Minimum enclosing and maximum excluding machine for pattern description and discrimination. In *ICPR '06: Proc. of the 18th International Conference on Pattern Recognition*, pp. 129–132, Washington, DC, USA. IEEE Computer Society.

Muñoz Marí, J., Bovolo, F., Gómez-Chova, L., Bruzzone, L., & Camp-Valls, G. (2010). Semi-Supervised One-Class Support Vector Machines for Classification of Remote Sensing Data. *IEEE Transactions on Geoscience and Remote Sensing*, *48*(8), 3188–3197.

Manevitz, L. M., & Yousef, M. (2002). One-class svms for document classification. *J. Mach. Learn. Res.*, *2*, 139–154.

Mao, C.-H., Lee, H.-M., Parikh, D., Chen, T., & Huang, S.-Y. (2009). Semi-supervised co-training and active learning based approach for multi-view intrusion detection. In *SAC '09: Proc. of the 2009 ACM symposium on Applied Computing*, pp. 2042–2048, New York, NY, USA. ACM.

Markou, M., & Singh, S. (2003a). Novelty detection: a review – part 1: statistical approaches. *Signal Processing*, *83*, 2481–2497.

Markou, M., & Singh, S. (2003b). Novelty detection: a review – part 2: neural network based approaches. *Signal Processing*, *83*, 2499–2521.

Maynor, K., Mookhey, K., Cervini, J. F. R., & Beaver, K. (2007). *Metasploit Toolkit*. Syngress.

Müller, K.-R., Mika, S., Rätsch, G., Tsuda, K., & Schölkopf, B. (2001). An introduction to kernel-based learning algorithms. *IEEE Neural Networks*, *12*(2), 181–201.

Onoda, T., Murata, H., & Yamada, S. (2006). One class classification methods based non-relevance feedback document retrieval. In *WI-IATW '06: Proc. of the 2006 IEEE/WIC/ACM international conference on Web Intelligence and Intelligent Agent Technology*, pp. 393–396, Washington, DC, USA. IEEE Computer Society.

Park, J., Kang, D., Kim, J., Kwok, J. T., & Tsang, I. W. (2007). SVDD-based pattern denoising. *Neural Computation*, *19*, 1919–1938.

Paxson, V. (1999). Bro: A System for Detecting Network Intruders in Real-Time. *Elsevier Computer Networks*, *31*(23-24), 2435–2463.

Pelleg, D., & Moore, A. (2004). Active learning for anomaly and rare-category detection. In *Proc. Advances in Neural Information Processing Systems*, pp. 1073–1080.

Perdisci, R., Ariu, D., Fogla, P., Giacinto, G., & Lee, W. (2009). McPAD: A multiple classifier system for accurate payload-based anomaly detection. *Computer Networks*, *5*(6), 864–881.

Rieck, K. (2009). *Machine Learning for Application-Layer Intrusion Detection*. Ph.D. thesis, Berlin Institute of Technology (TU Berlin).

Rieck, K., & Laskov, P. (2006). Detecting unknown network attacks using language models. In *Detection of Intrusions and Malware, and Vulnerability Assessment, Proc. of 3rd DIMVA Conference*, LNCS, pp. 74–90.

Rieck, K., & Laskov, P. (2007). Language models for detection of unknown attacks in network traffic. *Journal in Computer Virology*, 2(4), 243–256.

Rieck, K., & Laskov, P. (2008). Linear-time computation of similarity measures for sequential data. *Journal of Machine Learning Research*, 9(Jan), 23–48.

Rifkin, R. M., & Lippert, R. A. (2007). Value regularization and fenchel duality. *Journal of Machine Learning Research*, 8, 441–479.

Roesch, M. (1999). Snort: Lightweight intrusion detection for networks. In *Proc. of USENIX Large Installation System Administration Conference LISA*, pp. 229–238.

Salton, G., Wong, A., & Yang, C. (1975). A vector space model for automatic indexing. *Communications of the ACM*, 18(11), 613–620.

Schölkopf, B., & Smola, A. (2002). *Learning with Kernels*. MIT Press, Cambridge, MA.

Schölkopf, B., Platt, J. C., Shawe-Taylor, J., Smola, A. J., & Williamson, R. C. (2001). Estimating the support of a high-dimensional distribution. *Neural Computation*, 13(7), 1443–1471.

Shawe-Taylor, J., & Cristianini, N. (2004). *Kernel methods for pattern analysis*. Cambridge University Press.

Sindhwani, V., Niyogi, P., & Belkin, M. (2005). Beyond the point cloud: from transductive to semi-supervised learning. In *ICML*, Vol. 1, pp. 824–831. ACM.

Sonnenburg, S. (2008). *Machine Learning for Genomic Sequence Analysis*. Ph.D. thesis, Fraunhofer Institute FIRST. supervised by K.-R. Müller and G. Rätsch.

Stokes, J. W., & Platt, J. C. (2008). Aladin: Active learning of anomalies to detect intrusion. Tech. rep., Microsoft Research.

Stolfo, S. J., Apap, F., Eskin, E., Heller, K., Hershkop, S., Honig, A., & Svore, K. (2005). A comparative evaluation of two algorithms for windows registry anomaly detection. In *Journal of Computer Security*, pp. 659–693.

Tax, D. M. J. (2001). *One-class classification*. Ph.D. thesis, Technical University Delft.

Tax, D. M. J., & Duin, R. P. W. (2004). Support vector data description. *Machine Learning*, 54, 45–66.

Tong, S., & Koller, D. (2000). Support vector machine active learning with applications to text classification. In *Proc. of the Seventeenth International Conference on Machine Learning*, San Francisco, CA. Morgan Kaufmann.

Vapnik, V. (1998). *Statistical Learning Theory*. Wiley, New York.

Wang, J., Neskovic, P., & Cooper, L. N. (2005). Pattern classification via single spheres. In *Computer Science: Discovery Science (DS)*, Vol. 3735, pp. 241–252.

Wang, K., Parekh, J., & Stolfo, S. (2006). Anagram: A content anomaly detector resistant to mimicry attack. In *Recent Adances in Intrusion Detection (RAID)*, pp. 226–248.

Warmuth, M. K., Liao, J., Rätsch, G., Mathieson, M., Putta, S., & Lemmen, C. (2003). Active learning with support vector machines in the drug discovery process. *Journal of Chemical Information and Computer Sciences*, *43(2)*, 667–673.

Yuan, C., & Casasent, D. (2004). Pseudo relevance feedback with biased support vector machine. In *Proc. of the International Joint Conference on Neural Networks*.

Zhang, D., & Lee, W. S. (2005). A simple probabilistic approach to learning from positive and unlabeled examples. In *Proceedings of the 5th Annual UK Workshop on . . .* .